\documentclass{article}
\usepackage{spconf,amsmath,graphicx}

\usepackage{multirow}
\usepackage{amsmath,amssymb}
\usepackage{boldline}
\usepackage{bbm}
\usepackage{xcolor}
\title{Meta-Learned Feature Critics\\
for Domain Generalized Semantic Segmentation}

\name{Zu-Yun Shiau$^{1\ast}$ \qquad Wei-Wei Lin$^{2\ast}$ \qquad Ci-Siang Lin$^{1,3}$ \qquad Yu-Chiang Frank Wang$^{1,3}$\thanks{$\ast$ indicates equal contributions.}}

\address{$^{1}$Graduate Institute of Communication Engineering, National Taiwan University, Taiwan\\
$^{2}$Department of Electrical Engineering, National Taiwan University, Taiwan\\
$^{3}$ASUS Intelligent Cloud Services, Taiwan}

\begin{document}
%
\maketitle

\begin{abstract}

How to handle domain shifts when recognizing or segmenting visual data across domains has been studied by learning and vision communities. In this paper, we address domain generalized semantic segmentation, in which the segmentation model is trained on multiple source domains and is expected to generalize to unseen data domains. We propose a novel meta-learning scheme with feature disentanglement ability, which derives domain-invariant features for semantic segmentation with domain generalization guarantees. In particular, we introduce a class-specific feature critic module in our framework, enforcing the disentangled visual features with domain generalization guarantees. Finally, our quantitative results on benchmark datasets confirm the effectiveness and robustness of our proposed model, performing favorably against state-of-the-art domain adaptation and generalization methods in segmentation.

\end{abstract}

\begin{keywords}
domain generalization, semantic segmentation, meta-learning, deep learning, computer vision
\end{keywords}

\section{Introduction}
\label{sec:intro}

Deep learning has achieved great results in a wide variety of computer vision tasks. However, performance degrades when the trained models are applied to unknown domains \cite{chen2017no,vu2019advent} due to differences in training and testing domains. In this paper, we are concerned with the urban scene semantic segmentation task on unseen target domains.

Existing cross-domain segmentation approaches are trained on labeled source domain data, with unlabeled data jointly observed in the target domain. This setting is known as unsupervised domain adaptation (UDA)~\cite{chen2017no,vu2019advent,chang2019all,kim2020learning}. However, images of the desired target domain are not always available. Thus, recent research has turned to domain generalization methods, in which the model is trained solely on source domain data without access to target domain data. Recent works in this area have been mainly concerned with classification tasks, such as feature-based method~\cite{li2018domain} and meta-learning approaches~\cite{finn2017model,balaji2018metareg,li2019feature,li2019episodic}.

Domain generalization for semantic segmentation has seen less progress due to its task difficulty (pixel-wise classification) compared to classification. 
Recent works are model-based \cite{pan2018two}, embedding domain generalization capability into the model by modifying the model structure, or data-based \cite{yue2019domain}, augmenting source domain data with styles from additional datasets (e.g., ImageNet \cite{deng2009imagenet}). Although yielding positive results, their performance depend heavily on the style variance of the additional data introduced. Furthermore, while these methods provide style variation (e.g., textures, lighting, and colors), they do not guarantee generalization to unseen domains with structure difference caused by distinct camera view-points or cityscapes.

In this paper, we address the domain-generalized semantic segmentation problem. We propose a novel meta-learning scheme for learning domain-invariant visual features, which can be applied to unseen domains with segmentation guarantees. By advancing meta-learning strategies for \cite{lee2020drit++}, we apply feature disentanglement models to learn domain-invariant content features across multiple source domains. Moreover, inspired by \cite{li2019feature}, we introduce a class-specific feature critic module in the framework. This module additionally enforces the domain generalization capability of the learned features (e.g., invariant to domain shifts like viewpoint or cityscapes), thus realizing improved domain-generalized segmentation.

\begin{figure*}[t!]
	\centering
	\includegraphics[width=0.88\textwidth]{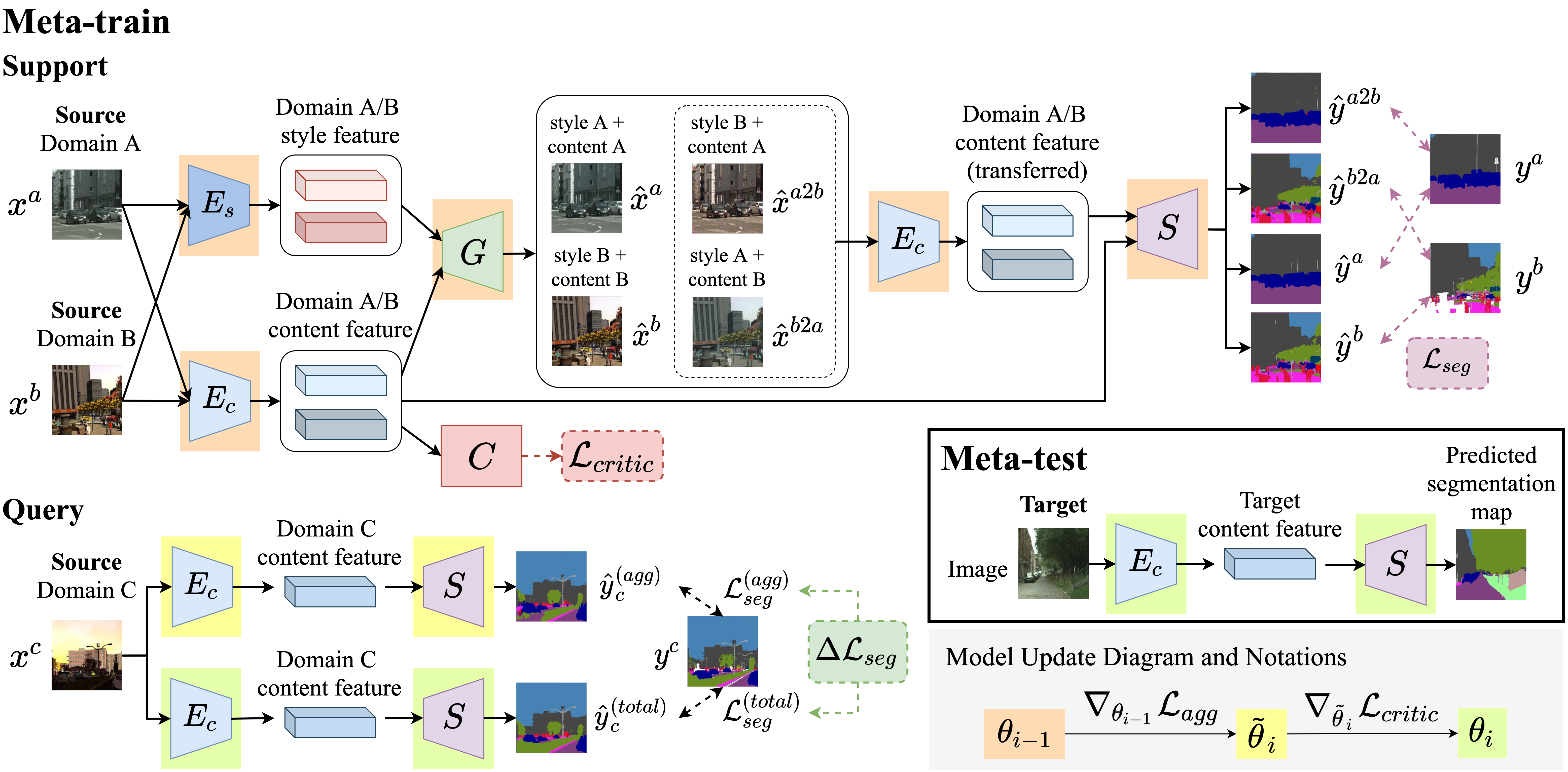}
 	\vspace{0mm}
    \caption{The architecture of our proposed model, which consists of a shared content encoder $E_c$, style encoder $E_s$, generator $G$, segmenter $S$, and feature critic module $C$. Note that different background colors denote models learned at different stages. While $\mathcal{L}_{agg}$ calculates disentanglement and segmentation losses $\mathcal{L}_{seg}$, $\mathcal{L}_{critic}$ particularly observes domain-generalized feature loss, which enforces $E_c$ to extract domain-invariant content features with domain-generalized segmentation guarantees.}
    \vspace{0mm}
	\label{fig:arch}
\end{figure*}

Our main contributions are listed as follows:
\begin{itemize}
\item We address the challenging task of domain generalized semantic segmentation, which performs pixel-wise label classification for input images observed in unseen domains.
\item In our proposed meta-learning scheme, we uniquely sample support and query sets across different source domains, enforcing the learned features for segmentation with domain generalization ability introduced.
\item To enforce domain generalization ability, we particularly introduce a class-specific feature critic module for semantic segmentation, which weighs disentangled visual features of each class properly, to achieve improved domain generalized segmentation.
\end{itemize}

%

\section{Proposed Method}
\label{sec:method}

\subsection{Problem definition and model overview}\label{ssec:definition}

We aim to learn a segmentation model from $N$ source domains $D=\{D_1, D_2, ..., D_N\}$ that would generalize to unseen target domains. Each domain contains a set of image and segmentation map pairs, i.e., $D_i=\{X^i, Y^i\}$. The objective is to predict $y^i$ based on the input image $x^i$. As illustrated in Fig.~\ref{fig:arch}, our proposed meta-learning framework includes: 1) domain-invariant feature disentanglement with domain generalization ability, and 2) learning class-specific feature critics with segmentation guarantees. The details of our proposed framework are presented in the following subsections. 

\subsection{Disentangling domain-invariant features for segmentation}\label{ssec:disen}
To realize domain-invariance property for segmentation, we require our model to extract and separate content features from the style ones. However, different from feature disentanglement techniques presented in~\cite{lee2020drit++}, we do not require explicit domain/style labels to realize our learning scheme. The feature disentanglement is performed by the guidance of the segmentation task as detailed below.

During meta-train, we sample an image pair across domains $D_a$ and $D_b$ (both from $D$), and deploy a shared content encoder $E_c$ and a shared style encoder $E_s$ for extracting content and style features of images ${x^a}\in D_a$, ${x^b}\in D_b$. We reconstruct images ${x_a}$, ${x_b}$ from their content and style features by a generator $G$ to yield:
\begin{equation*}
\begin{aligned}
\hat{x}^a=G(E_c(x^a), E_s(x^a)),\quad
\hat{x}^b=G(E_c(x^b), E_s(x^b)),
\end{aligned}
\end{equation*}
followed by the calculation of an L1 reconstruction loss $\mathcal{L}_{rec}$ between original and reconstructed images.


To separate the style and content of ${x^a}$ and ${x^b}$, we swap their styles by combining the content feature of one domain with the style feature of the other domain:
\begin{equation*}
\begin{aligned}
\hat{x}^{a2b}=G(E_c(x^a), E_s(x^b)),\quad
\hat{x}^{b2a}=G(E_c(x^b), E_s(x^a)).
\end{aligned}
\end{equation*}
We apply a perceptual loss $\mathcal{L}_{perc}$ calculated according to \cite{johnson2016perceptual} to enforce style and content constraints between original and style-transferred images. 

To alleviate style interference, we use only the content features for segmentation. The segmentation loss $\mathcal{L}_{seg}$ is given by cross-entropy loss between the ground truth and predicted segmentation maps $\hat{y}^{a}$, $\hat{y}^{b}$, $\hat{y}^{a2b}$, $\hat{y}^{b2a}$. Aggregating all three losses above yields the $\mathcal{L}_{agg}$ loss:
\begin{equation}\label{eq:agg}
\begin{aligned}
\mathcal{L}_{agg}=\lambda_{rec}\mathcal{L}_{rec}
                +\lambda_{perc}\mathcal{L}_{perc}
                +\lambda_{seg}\mathcal{L}_{seg}.
\end{aligned}
\end{equation}

\begin{figure}[t!]
	\centering
	\includegraphics[width=0.46\textwidth]{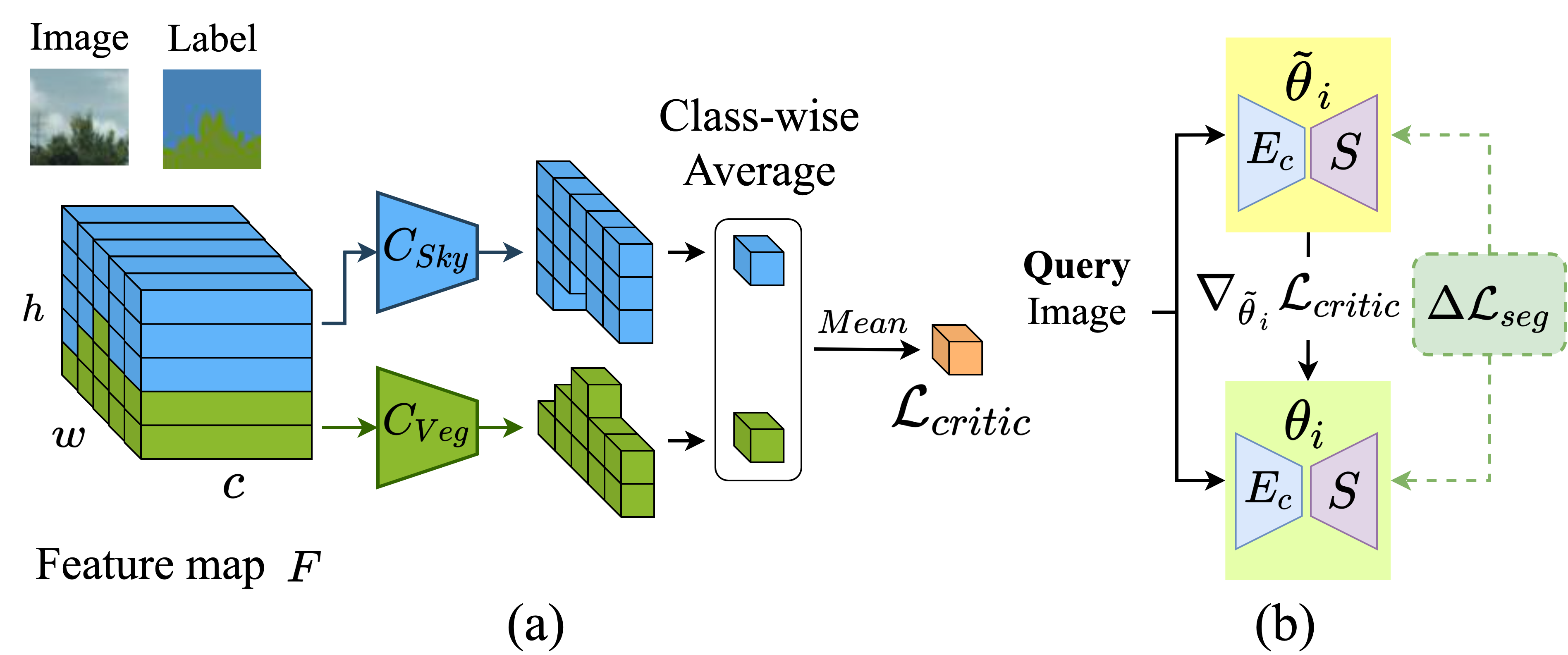}
 	\vspace{0mm}
    \caption{Illustration of our class-specific feature critics. (a) Example support set image with pixels associated with \textit{Sky} and \textit{Vegetation} ($C_{Sky}$ and $C_{Veg}$ are the corresponding feature critics). (b) $\mathcal{L}_{critic}$ updates models to enforce segmentation improvement on query images from a distinct domain.}
    \vspace{0mm}
	\label{fig:critic}
\end{figure}

\subsection{Class-specific feature critics for domain-generalized segmentation}\label{ssec:meta}
In addition to feature disentanglement on cross-domain data, we uniquely deploy a class-specific feature critic module $C$ in our proposed framework, introducing domain generalization ability to our segmentation model.

In previous classification work~\cite{li2019feature}, the feature critic transforms the entire learned cubic feature into a scalar output to describe the associated feature robustness. However, for semantic segmentation, each pixel obtains its prediction from a feature vector of dimension $c$ (channel size). Therefore, we propose to perform feature robustness evaluation on pixel-level feature vectors instead of the feature cube.

Our feature critic also differs from~\cite{li2019feature} in our exploitation of class information. The insight is that distinct classes rely on different features for segmentation, thus requiring distinct feature evaluation criteria. Therefore, we further propose to employ a class-specific feature critic module instead of a shared feature critic for pixels of distinct classes.

In our framework, the class-specific feature critic module consists of $K$ feature critics (MLP networks), where $K$ is the number of segmentation classes. We send all feature vectors of pixels of class $k$ into the corresponding feature critic $C_k$, as shown in Fig.~\ref{fig:critic}(a), where the sky pixels' feature vectors are sent into the sky feature critic. Then, we compute the class-wise average of each pixel's output from $C_k$:

\begin{equation}\label{eq:critic_k}
\begin{aligned}
\mathcal{L}_{critic}^{(k)}=\frac
{\sum_{h\in H, w\in W} C_{k}(f_{h, w})\mathbbm{1}[y_{h, w}=k]}
{\sum_{h\in H, w\in W} \mathbbm{1}[y_{h, w}=k]},
\end{aligned}
\end{equation}
where $f_{h, w}$ denotes the feature vector of one pixel, $y_{h, w}$ is its ground truth pixel-level label, and $\mathbbm{1}[\cdot]$ is an indicator function that outputs 1 when the condition is true or 0 otherwise. The domain-generalized feature loss $\mathcal{L}_{critic}$ is the mean of each class's feature loss $\mathcal{L}_{critic}^{(k)}$. As a feature robustness indicator, $\mathcal{L}_{critic}$ describes the domain generalization capabilities of $E_c$ and $S$. Learning of the feature critic module and how it encourages improved domain-generalized segmentation performances will be detailed in the following subsection.

\subsection{Meta-learning of domain-generalized features with class-specific feature critics}
We now discuss how we advance meta-learning strategies with multi-source data for domain-generalized segmentation.
The meta-train stage includes support and query stages in each iteration $i$, during which we sample two domains $D_a$, $D_b$ for support set, and one domain $D_c$ for query set from $D$.

For each sampled support set image pair, we perform feature disentanglement on $D_a$ and $D_b$, which yields the feature disentanglement losses $\mathcal{L}_{rec}$ and $\mathcal{L}_{perc}$, and segmentation loss $\mathcal{L}_{seg}$. The losses above constitute $\mathcal{L}_{agg}$~\eqref{eq:agg}.

We calculate intermediate parameters $\tilde{\theta}_i$ of all models by updating the previous iteration's parameters $\theta_{i-1}$ with $\mathcal{L}_{agg}$~\eqref{eq:theta_agg}. Then, all model parameters of iteration $i$, denoted by $\theta_i$, are computed by updating $\tilde{\theta}_i$ with the domain-generalized feature loss $\mathcal{L}_{critic}$~\eqref{eq:theta_total}. That is, 

\begin{equation}\label{eq:theta_agg}
\begin{aligned}
\tilde{\theta}_i=\theta_{i-1}-\alpha\nabla_{\theta_{i-1}}\mathcal{L}_{agg}
\end{aligned}
\end{equation}
\begin{equation}\label{eq:theta_total}
\begin{aligned}
\theta_{i}=\tilde{\theta}_i-\alpha\nabla_{\tilde{\theta}_i}\mathcal{L}_{critic}.
\end{aligned}
\end{equation}

At the same iteration and given a query image sampled from a distinct domain, we evaluate both sets of $E_c$ and $S$ parameters on the query set $D_c$. The resulting predicted segmentation maps are $\hat{y}_c^{(agg)}$ and $\hat{y}_c^{(total)}$, with segmentation losses $\mathcal{L}_{seg}^{(agg)}$ and $\mathcal{L}_{seg}^{(total)}$, based on the same ground truth $y_c$.
Since the domain-generalized feature loss serves to improve the generalization capability of $E_c$ and $S$, the model set updated with $\mathcal{L}_{critic}$ should perform better on the query set compared with the model set updated without $\mathcal{L}_{critic}$. We thus obtain the meta objective by maximizing prediction accuracy improvement between the two model sets:

\begin{equation}\label{eq:meta}
\begin{aligned}
\Delta\mathcal{L}_{seg}=\lambda_{meta}\times[-tanh(\mathcal{L}_{seg}^{(agg)}-\mathcal{L}_{seg}^{(total)})].
\end{aligned}
\end{equation}

We then update the class-specific feature critic module $C$ with the meta objective $\Delta\mathcal{L}_{seg}$. The domain-generalized feature loss enforces $E_c$ and $S$ to perform improved segmentation on cross-domain image data, while the meta-loss provides feedback for the class-specific feature critic module on its feature quality evaluation ability. As for inference in the meta-test stage, we directly take unseen target domain images with $E_c$ and $S$ for segmentation.


\section{Experiments}
\label{sec:experiments}
\subsection{Datasets and evaluation metrics}\label{ssec:dataset}

\begin{table}[!tp]
\centering
\caption{Quantitative comparisons with domain adaptation and generalization methods for segmentation in terms of mIoU. Note that both DANN~\cite{ganin2015unsupervised} and AdvEnt~\cite{vu2019advent} utilize target domain data to perform adaptation. C, G, S denote source datasets Cityscapes, GTA5, SYNTHIA, respectively, and XC is the target dataset Cross-City. P denotes additional data from \textit{Painter by Numbers}, which is utilized in LTIR~\cite{kim2020learning}.}
\vspace{1mm}
\resizebox{0.40\textwidth}{!}{
\begin{tabular}{c|c|c|c|c}
\multirow{2}{*}{Method} & \multicolumn{2}{c|}{Training Data} &
\multicolumn{2}{c}{mIoU} \\ \cline{2-5} & Source & Target & Source & Target
\\ \hlineB{2.5}
Baseline & C, G, S & - & 44.65 & 31.82 \\
DANN \cite{ganin2015unsupervised} & C, G, S & XC & 34.78 & 35.98 \\
AdvEnt \cite{vu2019advent} & C, G, S & XC & 39.86 & 36.53 \\
LTIR \cite{kim2020learning} & C, G, S, P & - & 40.34 & 36.42 \\
MMD-AAE \cite{li2018domain} & C, G, S & - & 52.54 & 39.20 \\
Ours & C, G, S & - & \textbf{65.66} & \textbf{44.59} \\
\end{tabular}}%
\vspace{0mm}
\label{tab:da_dg}
\end{table}

In our experiments, we consider Cityscapes \cite{cordts2016cityscapes}, GTA5 \cite{richter2016playing} and SYNTHIA \cite{ros2016synthia} as source domains, and Worldwide Road Scene Semantic Segmentation Dataset \cite{chen2017no} as target domain (referred to below as Cross-City). The source images are annotated pixel-wise by 19 semantic labels. Cross-City contains images of four cities, labeled by 13 major classes compatible with the source datasets. We take the mean Intersection-over-Union (mIoU) as the evaluation metric.

\subsection{Implementation details}\label{ssec:details}
Images are resized and cropped to $256\times256$ pixels during training, and $512\times1024$ pixels during testing. The models are trained for 450,000 iterations with batch size of 2. We use U-Net~\cite{ronneberger2015u} as backbone, the Adam optimizer for $E_s$ and $G$, and the SGD optimizer for $E_c$, $S$ and $C$. Finally, we set and fix hyper-parameters $\lambda_{rec} = 10$, $\lambda_{perc} = 1$, $\lambda_{seg} = 1$, $\lambda_{critic} = 0.1$ and $\lambda_{meta} = 5\times10^4$.

\subsection{Quantitative results and comparisons}\label{ssec:exp_comp}

For comparison purposes, we modify state-of-the-art domain adaptation and generalization approaches as follows. For DANN \cite{ganin2015unsupervised}, AdvEnt \cite{vu2019advent} and LTIR \cite{kim2020learning}, we conduct multi-source training. For MMD-AAE \cite{li2018domain}, we replace the classifier with a segmenter. The Baseline is trained on multi-source setting without adaptation methods. Table~\ref{tab:da_dg} compares the segmentation performances with U-Net as backbone. Note that we achieved the highest mIoU of \textbf{44.59} on the target set without viewing such data domains during training, nor observing additional data such as \textit{Painter by Numbers}.

\begin{table}[!tp]
\centering
\caption{Quantitative comparisons with additional domain generalization approaches. Following DRPC, the improvements of mIoU are the evaluation metrics.}
\vspace{1mm}
\resizebox{0.38\textwidth}{!}{
\begin{tabular}{c|c|c|c}
Backbone & Methods & mIoU & mIoU $\uparrow$ \\
\hlineB{2.5}
\multirow{4}{*}{ResNet50}
& Baseline$^1$ & 22.17 & \multirow{2}{*}{7.47} \\
& IBN-Net \cite{pan2018two} & 29.64 \\
\cline{2-4}
& Baseline$^1$ & 32.45 & \multirow{2}{*}{4.97} \\
& DRPC \cite{yue2019domain} & 37.42 \\
\hline
\multirow{2}{*}{Residual Blocks} & Baseline$^2$ & 31.51 & \multirow{2}{*}{\textbf{8.75}} \\
& Ours & 40.26 \\
\end{tabular}}
\vspace{0mm}
\label{tab:dg}
\end{table}

Table~\ref{tab:dg} further compares our results with recent domain generalization methods, IBN-Net \cite{pan2018two} and DRPC \cite{yue2019domain}. All baseline methods listed in this table indicate training on source domain data without generalization techniques (and testing on unseen target domain). Baseline$^1$ for IBN-Net and DRPC are both trained on GTA5 and tested on Cityscapes, while Baseline$^2$ for ours is trained on Cityscapes, GTA5 and SYNTHIA and tested on Cross-City. Our method achieved the mIoU improvement of \textbf{8.75}, which is above those reported by IBN-Net and DRPC. 

It is also worth noting that, the training set sizes of our method and DRPC are comparable (i.e., 36K (ours) vs. 41K (DRPC) images), and we do not require pre-training multiple CycleGAN-based models as DRPC does.

\begin{table}[!tp]
\centering
\caption{Ablation study of different components of our proposed model.}
\vspace{0mm}
\resizebox{0.39\textwidth}{!}{
\begin{tabular}{c|c|c|c|c}
\multirow{2}{*}{Method} & \multirow{2}{*}{Disent.} & \multirow{2}{*}{Meta}
    & \multicolumn{2}{c}{mIoU}
    \\ \cline{4-5} 
    & & & Source & Target
\\ \hlineB{2.5}
Baseline & - & - & 44.65 & 31.82 \\
Ours w/o Disent. & - & \checkmark & 60.18 & 34.89 \\
Ours w/o Meta & \checkmark & - & 61.52 & 43.26 \\
Ours & \checkmark & \checkmark & \textbf{65.66} & \textbf{44.59} \\
\end{tabular}}
\vspace{0mm}
\label{tab:abl}
\end{table}


\subsection{Ablation study}\label{ssec:exp_abla}
To assess the contributions of different parts of our method, we conduct an ablation study as shown in Table~\ref{tab:abl}. \textit{Baseline} trains with all source domains without adaptation methods. \textit{Ours w/o Disent.} denotes meta-learning a class-specific feature critic module without disentanglement. \textit{Ours w/o Meta} denotes performing disentanglement without meta-learning. Finally, \textit{Ours} is the full version of our proposed method. As shown in the table, meta-learning of our feature critic module improved the baseline by \textbf{3.07}, and disentanglement improved the baseline by \textbf{11.44}, confirming the effectiveness of conducting segmentation on content features only. \textit{Ours} improved upon \textit{Ours w/o Disent.} and \textit{Ours w/o Meta}, demonstrating that all parts contribute to generalization ability.

\section{Conclusions}
\label{sec:conclusions}
In this paper, we propose a meta-learning based model for domain generalized semantic segmentation. By utilizing multiple source domains and meta-learning strategies, our proposed model jointly performs feature disentanglement and learns class-specific feature critics. Thus, domain-invariant content features can be extracted for domain-generalized segmentation. We study the effectiveness of different parts of our method, and demonstrate that our method performed favorably against state-of-the-art domain generalization or adaptation methods.

\section{Acknowledgement}
This work is supported in part by the Ministry of Science and Technology of Taiwan under grant MOST 110-2634-F-002-036. We also thank the National Center for High-performance Computing (NCHC) for providing computational and storage resources.

\bibliographystyle{IEEEbib}
\bibliography{strings,refs}

\end{document}